\documentclass[11pt,a4paper]{article}
\usepackage[hyperref]{emnlp2020}
\usepackage{times}
\usepackage{latexsym}

\usepackage{amsmath,amsfonts,bm}

\def\eqref#1{equation~(\ref{#1})}

\def\1{\bm{1}}

\def\vv{{\bm{v}}}

\DeclareMathAlphabet{\mathsfit}{\encodingdefault}{\sfdefault}{m}{sl}
\SetMathAlphabet{\mathsfit}{bold}{\encodingdefault}{\sfdefault}{bx}{n}

\DeclareMathOperator*{\argmax}{arg\,max}
\DeclareMathOperator*{\argmin}{arg\,min}

\usepackage{microtype}

\usepackage{graphicx}
\usepackage{xcolor}
\usepackage{multirow}

\DeclareMathOperator*{\LM}{TLM}

\DeclareMathOperator*{\attention}{attention}

\usepackage{amssymb}
\usepackage{graphicx}
\usepackage{subcaption}

\usepackage{ctable}

\newcommand{\Expect}{{\rm I\kern-.3em E}}

\DeclareMathOperator*{\MLP}{MLP}

\DeclareMathOperator*{\product}{VKP}

\newcommand{\x}{\boldsymbol{x}}
\newcommand{\y}{\boldsymbol{y}}
\newcommand{\xspace}{\mathcal{X}}
\newcommand{\yspace}{\mathcal{Y}}
\newcommand{\relyspace}{\mathcal{Y}_{R}}
\newcommand{\cost}{\bigtriangleup}
\newcommand{\infnet}{\mathbf{A}_{\Psi}}
\newcommand{\canet}{\mathbf{F}_{\Phi}}

\newcommand{\ltok}{\ell_{\mathrm{token}}}
\newcommand{\pvect}{\mathbf{v}}

\usepackage{microtype}

\aclfinalcopy 

\title{An Exploration of Arbitrary-Order Sequence Labeling\\
via Energy-Based Inference Networks}

\author{Lifu Tu$^1$ \Thanks{ Equal contribution.} \ \ \ \ \ \   Tianyu Liu$^2$ \footnotemark[1] \ \ \ \ \ \ \ Kevin Gimpel$^1$  \\
 $^1$ Toyota Technological Institute at Chicago, IL, USA \\
 $^2$ Institute of Computational Linguistics, Peking University, China \\
  \texttt{\{lifu, kgimpel\}@ttic.edu,} \ \texttt{tianyu0421@pku.edu.cn} \\}

\date{}

\begin{document}
\maketitle
\begin{abstract}
Many tasks in natural language processing involve predicting structured outputs, e.g., sequence labeling, semantic role labeling, parsing, and machine translation. Researchers are increasingly applying deep representation learning to these problems, but the structured component of these approaches is usually quite simplistic. In this work, we propose several high-order energy terms to capture complex dependencies among labels in sequence labeling, including several that consider the entire label sequence. We use neural parameterizations for these energy terms, drawing from convolutional, recurrent, and self-attention networks. We use the framework of learning energy-based inference networks~\citep{tu-18} for dealing with the difficulties of training and inference with such models. We empirically demonstrate that this approach achieves substantial improvement using a variety of high-order energy terms on four sequence labeling tasks, while having the same decoding speed as simple, local classifiers. 
We also find high-order energies to help in noisy data conditions.\footnote{Code 
is available at \url{https://github.com/tyliupku/Arbitrary-Order-Infnet}}
\end{abstract}

\newcommand{\fix}{\marginpar{FIX}}
\newcommand{\new}{\marginpar{NEW}}

\section{Introduction}

Conditional random fields (CRFs; \citealp{Lafferty:2001:CRF:645530.655813}) have been shown to perform well in various sequence labeling tasks. Recent work uses rich neural network architectures to define the ``unary'' potentials, i.e., terms that only consider a single position's label at a time~\citep{2011:NLP-sm,LampleBSKD16,ma-hovy:2016:P16-1,strubell-etal-2018-linguistically}. However, ``binary'' potentials, which consider pairs of adjacent labels, are usually quite simple and may consist solely of a parameter or parameter vector for each unique label transition. Models with unary and binary potentials are generally referred to as ``first order'' models. 

A major challenge with CRFs is the complexity
of training and inference, which are
quadratic in the number of output labels for first order models and grow exponentially when higher order dependencies are considered. This explains why the most common type of CRF used in practice is a first order model, also referred to as a ``linear chain'' CRF. 

One promising alternative to CRFs is structured prediction energy networks (SPENs; \citealp{belanger2016structured}), which use deep neural networks to parameterize arbitrary potential functions for structured prediction. While SPENs also pose challenges for learning and inference, \citet{tu-18} proposed a way to train SPENs jointly with ``inference networks'', neural networks trained to approximate structured $\argmax$ inference. 

In this paper, we leverage the frameworks of SPENs and inference networks to explore high-order energy functions for sequence labeling. Naively instantiating high-order energy terms can lead to a very large number of parameters to learn, so we instead develop concise neural parameterizations for high-order terms. In particular, we draw from vectorized Kronecker products, convolutional networks, recurrent networks, and self-attention. 
We also consider ``skip-chain'' connections~\citep{sutton04skip} with various skip distances and ways of reducing their total parameter count for increased learnability. 

Our experimental results on four sequence labeling tasks show that a range of high-order energy functions can yield performance improvements. While the optimal energy function varies by task, we find strong performance from skip-chain terms with short skip distances, convolutional networks with filters that consider label trigrams, and recurrent networks and self-attention networks that consider large subsequences of labels.

We also demonstrate that modeling high-order dependencies can lead to significant performance improvements in the setting of noisy training and test sets. 
Visualizations of the high-order energies show various methods capture intuitive structured dependencies among output labels. 

Throughout, we use inference networks that share the same architecture as unstructured classifiers for sequence labeling, so test time inference speeds are unchanged between local models and our method. 
Enlarging the inference network architecture by adding one layer leads consistently to better results, rivaling or improving over a BiLSTM-CRF baseline, 
suggesting that training efficient inference networks with high-order energy terms can make up for errors arising from approximate inference.
While we focus on sequence labeling in this paper, our results show the potential of developing high-order structured models for other NLP tasks in the future.

\section{Background} 
\label{sec:background}

\subsection{Structured Energy-Based Learning}

We denote the input space by $\xspace$. For an input $\x\in\xspace$, we denote the structured  output space by $\yspace(\x)$. The entire space of structured outputs is denoted $\yspace = \cup_{\x\in\xspace} \yspace(\x)$. 
We define an \textbf{energy function}~\citep{lecun-06, belanger2016structured}  $E_{\Theta}$ parameterized by $\Theta$ that 
computes a scalar energy for an input/output pair: $E_{\Theta} : \xspace \times \yspace \rightarrow \mathbb{R}$. 
At test time, for a given input $\x$, prediction is done by choosing the output with lowest energy:
\begin{align}\label{eq:inf}
\textstyle{\hat{\y} = \argmin_{\y\in\yspace(\x)}E_{\Theta}(\x, \y)}
\end{align}

\subsection{Inference Networks}

\paragraph{Inference.}  Solving \eqref{eq:inf} requires combinatorial algorithms because $\yspace$ is a structured, discrete space. This becomes intractable when $E_{\Theta}$ does not decompose into a sum over small ``parts'' of $\y$. 
\citet{belanger2016structured} relax this problem by allowing the discrete vector $\y$ to be continuous.  Let $\relyspace$ denote the \textbf{relaxed output space}. 
They solve the relaxed problem by using gradient descent to iteratively minimize the energy with respect to $\y$.

\citet{tu-18} propose an alternative that replaces gradient descent  
with a neural network trained to do inference, i.e., to mimic the function performed in \eqref{eq:inf}. This ``inference network'' $\infnet : \xspace \rightarrow \relyspace$ is parameterized by $\Psi$ and trained with the goal that
\begin{align}\label{eqn:original-psi}
\infnet(\x) \approx \argmin_{\y\in\relyspace(\x)}E_\Theta(\x, \y)
\end{align}
\citet{tu-gimpel-2019-benchmarking} show that inference networks achieve a better speed/accuracy/search error trade-off than gradient descent given pretrained energy functions. 

\paragraph{Joint training of energy functions and inference networks.}

\citet{belanger2016structured} proposed a structured hinge loss for learning the energy function parameters $\Theta$, using gradient descent for the ``cost-augmented'' inference step required during learning. \citet{tu-18} replaced the cost-augmented inference step in the structured hinge loss with training of a ``cost-augmented inference network'' $\canet(\x)$ trained with the following goal: 
$$\canet(\x) \approx \argmin_{\y\in\relyspace(\x)}\,  (E_\Theta(\x, \y)-\cost(\y, \y^\ast))$$
\noindent where $\cost$ is a structured cost function that computes the distance between its two arguments.  The new optimization objective becomes: 
\begin{align}
\small
\min_{\Theta} \max_{\Phi} \sum_{\langle \x, \y\rangle\in\mathcal{D}} & [ \cost(\canet(\x), \y) \nonumber \\ & -  E_{\Theta}(\x,\canet(\x))  + E_{\Theta}(\x, \y) ]_{+} \nonumber
\end{align}
\noindent where $\mathcal{D}$ is the set of training pairs and $[h]_+ = \max(0,h)$. 
\citet{tu-18} 
alternatively optimized $\Theta$ and $\Phi$, which is similar to training in generative adversarial networks~\citep{goodfellow2014generative}. 

\subsection{An Objective for Joint Learning of Inference Networks}
\label{sec:joint}

One challenge with the optimization problem above is that it still requires training an inference network $\infnet$ for test-time prediction. \citet{tu2019improving} proposed a ``compound'' objective that avoids this by training two inference networks jointly (with shared parameters), $\canet$ for cost-augmented inference and $\infnet$ for test-time inference:
\begin{align}
& \!\min_{\Theta} \max_{\Phi, \Psi} \!\!\!\!\sum_{\langle \x, \y\rangle\in\mathcal{D}} \nonumber \\ 
& \underbrace{[\cost(\canet(\x), \y) \!-\! E_{\Theta}(\x,\canet(\x)) \!+\! E_{\Theta}(\x, \y)]_{+}}_{\text{margin-rescaled hinge loss}} \nonumber \\
& + 
\lambda \underbrace{\left[- E_{\Theta}(\x,\infnet(\x)) \!+\! E_{\Theta}(\x, \y)\right]_{+}}_{\text{perceptron loss}} \nonumber
\end{align}
As indicated, this loss can be viewed as the sum of the margin-rescaled and perceptron losses. 
$\Theta$, $\Phi$, and $\Psi$ are alternatively optimized. The objective for the energy function parameters $\Theta$ is:
\begin{align}
&\hat{\Theta} \gets 
\argmin_{\Theta} \nonumber \\ 
&\big[\!\cost\!(\canet(\x), \y) - E_{\Theta}(\x,\canet(\x)) + E_{\Theta}(\x, \y)\big]_{+} \nonumber \\ 
& + \lambda \big[-\!E_{\Theta}(\x,\infnet(\x)) + E_{\Theta}(\x, \y)\big]_{+} \nonumber
\end{align}
The objective for the other parameters is:
\begin{align}
\hat{\Psi}, \hat{\Phi} & \gets  \argmax_{\Psi,\Phi} \cost(\canet(\x), \y) - E_{\Theta}(\x,\canet(\x)) \nonumber \\
& - \lambda E_{\Theta}(\x,\infnet(\x)) - \tau \ltok(\y, \infnet(\x))
\nonumber
\end{align}
\noindent where $\ltok$ is a supervised token-level loss which is added to aid in training inference networks. In this paper, we use the standard cross entropy summed over all positions. Like \citet{tu2019improving}, we drop the zero truncation ($\max(0,.)$) when updating the inference network parameters to improve stability during training, which also lets us remove the terms that do not have inference networks. We use two independent networks but with the same architecture for the two inference networks.

\section{Energy Functions}
\label{sec:energy}
Our experiments in this paper consider sequence labeling tasks, so the input $\x$ is a length-$T$ sequence of tokens where $x_t$ denotes the token at position $t$. The output $\y$ is a sequence of labels also of length $T$. 
We use $\y_t$ to denote the output label at position $t$, where $\y_t$ is a vector of length $L$ (the number of labels in the label set) and where $y_{t,j}$ is the $j$th entry of the vector $\y_t$. In the original output space $\yspace(\x)$, $y_{t,j}$ is 1 for a single $j$ and 0 for all others. In the relaxed output space $\relyspace(\x)$, $y_{t,j}$  can be interpreted as the probability of the $t$th position being labeled with label $j$. 
We use the following energy: $E_{\Theta}(\x, \y) = $
\begin{align}
-\Bigg(\sum_{t=1}^T  \sum_{j=1}^L y_{t,j}\! \left(U_j^\top b(\x,t)\right) 
+  E_{W}( \y) \Bigg) 
\label{eqn:energy}
\end{align}
where $U_{j}\in\mathbb{R}^d$ is a parameter vector for label $j$  
and $E_{W}( \y)$ is a structured energy term parameterized by parameters $W$. In a linear chain CRF, $W$ is a transition matrix for scoring two adjacent labels. Different instantiations of $E_{W}$ will be detailed in the sections below. 
Also, $b(\x,t)\in\mathbb{R}^d$ denotes the ``input feature vector'' for position $t$. 
We define it to be the $d$-dimensional BiLSTM~\citep{Hochreiter:1997:LSM:1246443.1246450} hidden vector at $t$.
The full set of energy parameters $\Theta$ includes the $U_j$ vectors, $W$, and the parameters of the BiLSTM. 

The above energy functions are trained with the objective in Section~\ref{sec:joint}. Table~\ref{tab:complexity} shows the training and test-time inference requirements 
of our method compared to previous methods. For different formulations of the energy function, the inference network architecture is the same (e.g., BiLSTM). So the inference complexity is the same as the standard neural approaches that do not use structured prediction, which is linear in the label set size. However, even for the first order model (linear-chain CRF), the time complexity is quadratic in the label set size. The time complexity of higher-order CRFs grows exponentially with the order. 

\begin{table*}[t]
    \setlength{\tabcolsep}{6pt}
    \small
    \centering
    \begin{tabular}{lc|c|c||c|c|}
\cline{3-6}
&& \multicolumn{2}{c||}{Training}
& \multicolumn{2}{c|}{Inference}
\\
&& \multicolumn{1}{c}{Time} & \multicolumn{1}{c||}{Number of Parameters}  & \multicolumn{1}{c}{Time} & \multicolumn{1}{c|}{Number of Parameters} \\ 
\hline
\multicolumn{2}{|c|}{BiLSTM} &  $\mathcal{O}(T*L)$ & $\mathcal{O}(|\Psi|)$  & $\mathcal{O}(T*L)$ & $\mathcal{O}(|\Psi|)$ \\
\multicolumn{2}{|c|}{CRF} & $\mathcal{O}(T*L^2)$   & $\mathcal{O}(|\Theta|)$  & $\mathcal{O}(T*L^2)$ & $\mathcal{O}(|\Theta|)$ \\
\hline \hline
\multicolumn{2}{|l|}{Energy-Based Inference Networks} & $\mathcal{O}(T*L)$  & $\mathcal{O}(|\Psi| + |\Phi| + |\Theta|)$  & $\mathcal{O}(T*L)$ & $\mathcal{O}(|\Psi|)$\\
\hline
    \end{tabular}
    \caption{Time complexity and number of parameters of different methods during training and inference, where $T$ is the sequence length, $L$ is the label set size, $\Theta$ are the parameters of energy function, and $\Phi, \Psi$ are the parameters of two energy-based inference networks. For arbitrary-order energy functions or different parameterizations, the size of $\Theta$ can be different.
    }
    \label{tab:complexity}
\end{table*}

\subsection{Linear Chain Energies} 

Our first choice for a structured energy term is relaxed linear chain energy defined for sequence labeling by \citet{tu-18}: 
\begin{align}
E_{W}( \y) = \sum_{t=1}^T \y_{t-1}^\top W \y_{t}  \nonumber
\end{align}
Where $W_i\in\mathbb{R}^{L \times L}$ is the transition matrix, which is used to score the pair of adjacent labels. If this linear chain energy is the only structured energy term in use, exact inference can be performed efficiently using the Viterbi algorithm. 

\subsection{Skip-Chain Energies}
\label{sec:skipchain}
We also consider an energy inspired by ``skip-chain'' conditional random fields~\citep{sutton04skip}. In addition to consecutive labels, this energy also considers pairs of labels appearing in a given window size $M+1$:
\begin{align}
E_{W}( \y) = \sum_{t=1}^T \sum_{i=1}^M \y_{t-i}^\top W_i \y_{t}  \nonumber
\end{align}
where each $W_i\in\mathbb{R}^{L \times L}$ and the max window size $M$ is a hyperparameter. While linear chain energies allow efficient exact inference, using skip-chain energies causes exact inference to require time exponential in the size of $M$.

\subsection{High-Order Energies}
\label{sec:hoEnergy}
We also consider $M$th-order energy terms. We use the function $F$ to score the $M+1$ consecutive labels $\y_{t-M}, \dots, \y_{t}$, then sum over positions:  
\begin{align}
E_{W}( \y) = \sum_{t=M}^T F(\y_{t-M},\dots,\y_t) 
\label{eqn:high-order}
\end{align}
\noindent We consider several different ways to define the function $F$, detailed below.
\paragraph{Vectorized Kronecker Product (VKP):} 
A naive way to parameterize a high-order energy term would involve using a parameter tensor $W \in\mathbb{R}^{L^{M+1}}$ with an entry for each possible label sequence of length $M+1$. To avoid this exponentially-large number of parameters, we define a more efficient parameterization as follows. 
We first define a label embedding lookup table $\in \mathbb{R}^{L \times n_l}$ and denote the embedding for label $j$ by $e_j$. We consider $M=2$ as an example. Then, for a tensor $W \in\mathbb{R}^{L\times L\times L}$, its value $W_{i,j,k}$ at indices $(i,j,k)$ is calculated as 
$$
\pvect^\top \mathrm{LayerNorm}([e_i;e_j;e_k]+\MLP([e_i;e_j;e_k]))$$
\noindent where $\pvect \in \mathbb{R}^{(M+1) n_l}$ is a parameter vector and $;$ denotes vector concatenation. $\MLP$ expects and returns vectors of dimension ${(M+1) \times n_l}$ and is parameterized as a multilayer perceptron. Then, the energy is computed:
\begin{align}
F(\y_{t-M},\dots,\y_t) = \product(\y_{t-M}, \dots, \y_{t-1})W \y_t
\nonumber 
\end{align}
\noindent where $W$ is reshaped as $\in\mathbb{R}^{L^M \times L}$. 
The operator $\product$ is somewhat similar to the Kronecker product of the $k$ vectors $\vv_{1}, \dots, \vv_{k}$\footnote{There are some work~\citep{lei-etal-2014-low,NIPS2014_5323, yu-etal-2016-embedding} that use Kronecker product for higher order feature combinations with low-rank tensors. Here we use this form to express the computation when scoring the consecutive labels. }. However it will return a vector, not a tensor: 
\begin{align}
    & \product(\vv_{1}, \dots, \vv_{k}) = \nonumber \\
    &\begin{cases}
     \vv_{1} & k=1 \\
     \mathbf{vec}(\vv_{1} \vv_{2}^\top) & k=2 \\
     \mathbf{vec}(\product(\vv_1,\dots, \vv_{k-1}) \vv_{k}^\top) & k > 2
    \end{cases} \nonumber
\end{align}
\noindent Where $\mathbf{vec}$ is the operation that vectorizes a tensor into a (column) vector.

\paragraph{CNN:} Convolutional neural networks (CNN)  are frequently used in NLP to extract features based on words or characters~\citep{2011:NLP-sm,kim-2014-convolutional}. We apply CNN filters over the sequence of $M+1$ consecutive labels. 
The $F$ function is computed as follows:
\begin{align}
&F(\y_{t-M},\dots,\y_t) = \sum_n f_{n}(\y_{t-M},\dots,\y_t) \nonumber \\
&f_{n}(\y_{t-M},\dots,\y_t) = g(W_n [\y_{t-M};
...;\y_{t} ] +\mathbf{b}_n)  \nonumber
\end{align}
\noindent where $g$ is a ReLU nonlinearity and the vector $W_n \in\mathbb{R}^{L(M+1)}$ and scalar $b_n \in\mathbb{R}$ are the parameters for filter $n$. 
The filter size of all filters is the same as the window size, namely, $M+1$. The $F$ function sums over all CNN filters. When viewing this high-order energy as a CNN, we can think of the summation in Eq.~\ref{eqn:high-order} as corresponding to sum pooling over time of the feature map outputs.

\paragraph{Tag Language Model (TLM):} \citet{tu-18} defined an energy term based on a pretrained ``tag language model'', which computes the probability of an entire sequence of labels. We also use a TLM, scoring a sequence of $M+1$ consecutive labels in a way similar to \citet{tu-18}; however, the parameters of the TLM are trained in our setting:
\begin{align}
& F(\y_{t-M},\dots,\y_t) =  \nonumber \\
& - \! \sum_{t'=t-M+1}^{t} \! \y_{t'}^\top \log(\LM(\langle \y_{t-M},...,\y_{t'-1}\rangle) ) \nonumber
\label{eq:tlm}
\end{align}
\noindent where $\LM(\langle \y_{t-M},...,y_{t'-1}\rangle)$ returns the softmax distribution over tags at position $t'$ (under the  tag language model) given the preceding tag vectors. When each $y_{t'}$ is a one-hot vector, this energy reduces to the negative log-likelihood of the tag sequence specified by $\y_{t-M},\dots, \y_t$.

\paragraph{Self-Attention (S-Att):} We adopt the multi-head self-attention formulation from \citet{NIPS2017_7181}. Given a matrix of the $M+1$ consecutive labels $Q=K=V= [\y_{t-M};\dots;\y_{t} ] \in \mathbb{R}^{(M+1) \times L}$: 
\begin{align}
    H = \attention(Q, K, V) \nonumber \\
    F(\y_{t-M},\dots,\y_t) = \sum H \nonumber
\end{align}
\noindent where $\attention$ is the general attention mechanism: the weighted sum of the value vectors $V$ using query vectors $Q$ and key vectors $K$ \citep{NIPS2017_7181}. The energy on the $M+1$ consecutive labels is defined as the sum of entries in the feature map $H \in \mathbb{R}^{L\times (M+1)}$ after the self-attention transformation.

\subsection{Fully-Connected Energies}
We can simulate a ``fully-connected'' energy function by setting a very large value for $M$ in the skip-chain energy (Section \ref{sec:skipchain}). For efficiency and learnability, we
use a low-rank parameterization for the many translation matrices $W_i$ that will result from increasing $M$. We first define a matrix $S\in \mathbb{R}^{L \times d}$ that all $W_i$ will use. Each $i$ has a learned parameter matrix $D_i \in \mathbb{R}^{L \times d} $ and together $S$ and $D_i$ are used to compute $W_i$: 
\begin{align}
    W_i = S D_i^\top \nonumber
\end{align}
\noindent where $d$ is a tunable hyperparameter that affects the number of learnable parameters.

\section{Related Work}

Linear chain CRFs \citep{Lafferty:2001:CRF:645530.655813},  which consider dependencies between at most two adjacent labels or segments, are commonly used in practice \citep{NIPS2004_2648,LampleBSKD16,ma-hovy:2016:P16-1}. 

There have been several efforts in developing efficient algorithms for handling higher-order CRFs. 
\citet{QianJZHW09} developed an efficient decoding algorithm under the assumption that all  high-order features have non-negative weights. Some work has shown that high-order CRFs can be handled relatively efficiently if particular patterns of sparsity are assumed~\citep{NIPS2009_3815,JMLR:v15:cuong14a}. \citet{mueller-etal-2013-efficient} proposed an approximate CRF using coarse-to-fine decoding and early updating. 
Loopy belief propagation~\citep{MurphyWJ99} has been used for  approximate inference in high-order CRFs, such as skip-chain CRFs~\citep{sutton04skip}, which form the inspiration for one category of energy function in this paper. .

CRFs are typically trained by maximizing conditional log-likelihood. Even assuming that the graph structure underlying the CRF admits tractable inference, it is still time-consuming to compute the partition function. Margin-based methods have been proposed~\citep{m3,ssvm} to avoid the summation over all possible outputs. Similar losses are used when training SPENs~\citep{belanger2016structured,End-to-EndSPEN}, including in this paper. .  
The energy-based inference network learning framework has been used for multi-label classification \citep{tu-18}, non-autoregressive machine translation \citep{tu-2020-nat}, and previously for sequence labeling \citep{tu-gimpel-2019-benchmarking}.

Moving beyond CRFs and sequence labeling, there has been a great deal of work in the NLP community in designing non-local features,
often combined with the development of approximate algorithms to incorporate them during inference. 
These include $n$-best reranking \citep{och-etal-2004-smorgasbord}, 
beam search \citep{Lowerre:1976:HSR:907741}, 
loopy belief propagation \citep{sutton04skip,smith-eisner-2008-dependency}, 
Gibbs sampling \citep{finkel-etal-2005-incorporating}, 
stacked learning \citep{DBLP:conf/ijcai/CohenC05,krishnan-manning-2006-effective}, 
sequential Monte Carlo algorithms~\citep{yang-eisenstein-2013-log}, 
dynamic programming approximations like cube pruning~\citep{chiang-2007-hierarchical,huang-chiang-2007-forest},
dual decomposition \citep{rush-etal-2010-dual,martins-etal-2011-dual}, 
and methods based on black-box optimization like integer linear programming~\citep{roth-yih-2004-linear}. These methods are often developed or applied with particular types of non-local energy terms in mind. By contrast, here we find that the framework of SPEN learning with inference networks can support a wide range of high-order energies for sequence labeling.

\section{Experimental Setup}
We perform experiments on four tasks: Twitter part-of-speech tagging (POS), named entity recognition (NER), CCG supertagging (CCG), and semantic role labeling (SRL).

\subsection{Datasets}
\paragraph{POS.}
We use the annotated data from \citet{gimpel-11a} and \citet{owoputi-EtAl:2013:NAACL-HLT} which contains
25 POS tags.
We use the 100-dimensional skip-gram embeddings from \citet{tu-17-long} which were trained on a dataset of 56 million English tweets using  \texttt{word2vec}~\citep{mikolov2013distributed}. The evaluation metric is tagging accuracy.

\paragraph{NER.}
We use the CoNLL 2003 English data \citep{tjong-kim-sang-de-meulder-2003-introduction}. 
We use the BIOES tagging scheme, so there are 17 labels. 
We use 100-dimensional pretrained GloVe~\citep{pennington2014glove} embeddings.  
The task is evaluated with micro-averaged F1 score.  

\paragraph{CCG.}
We use the standard splits from CCGbank~\citep{hockenmaier-steedman-2002-acquiring}.
We only keep sentences with length less than 50 in the original training data during training. 
We use only the 400 most frequent labels.
The training data contains 1,284 unique labels, but  because the label distribution has a long tail,  we use only the 400 most frequent labels, replacing the others by a special tag $*$. The percentages of $*$ in train/development/test are 0.25/0.23/0.23$\%$.  When the gold standard tag is $*$, the prediction is always evaluated as incorrect. We use the same GloVe embeddings as in NER. . 
The task is evaluated with per-token accuracy.

\paragraph{SRL.}
We use the standard split from CoNLL 2005 \citep{carreras-marquez-2005-introduction}. The gold predicates are provided as part of the input. We use the official evaluation script from the CoNLL 2005 shared task for  evaluation. 
We again use the same GloVe embeddings as in NER. 
To form the inputs to our models, an embedding of a binary feature indicating whether the word is the given predicate is concatenated to the word embedding.\footnote{Our SRL baseline  is most similar to \citet{zhou-xu-2015-end}, though there are some differences. 
We use GloVe embeddings while they train word embeddings on Wikipedia. We both use the same predicate context features.} 

\subsection{Training}
\paragraph{Local Classifiers.}
We consider local baselines that use a BiLSTM trained with the local loss $\ltok$. For POS, NER and CCG, we use a 1-layer BiLSTM with hidden size 100, and the word embeddings are fixed during training. For SRL, we use a 4-layer BiLSTM with hidden size 300 and the word embeddings are fine-tuned. 

\paragraph{BiLSTM-CRF.}
We also train BiLSTM-CRF models with the standard conditional log-likelihood objective. A 1-layer BiLSTM with hidden size 100 is used for extracting input features. The CRF part uses a linear chain energy with a single tag transition parameter matrix. 
We do early
stopping based on development sets. The usual dynamic programming algorithms are used for training and inference, e.g., the Viterbi algorithm is used for inference. The same pretrained word embeddings as for the local classifiers are used.

\paragraph{Inference Networks.}
When defining architectures for the inference networks, we use the same architectures as the local classifiers. However, the objective of the inference networks is different, which is shown in Section~\ref{sec:joint}. $\lambda=1$ and $\tau=1$ are used for training. We do early
stopping based on the development set.

\paragraph{Energy Terms.}
The unary terms are parameterized using a one-layer BiLSTM with hidden size 100. 
For the structured energy terms, the $\product$ operation uses $n_l=20$, the number of CNN filters is 50, and the tag language model is a 1-layer LSTM with hidden size 100. For the fully-connected energy, $d=20$ for the approximation of the transition matrix and $M=20$ for the approximation of the fully-connected energies.

\paragraph{Hyperparameters.}
For the inference network training, the batch size is 100. We update the energy function parameters using the Adam optimizer~\citep{adam} with learning rate 0.001. For POS, NER, and CCG, we train the inference networks parameter with stochastic gradient descent with momentum as the optimizer. The learning rate is 0.005 and the momentum is 0.9. For SRL, we train the inference networks using Adam with learning rate 0.001. 

\section{Results}

\paragraph{Parameterizations for High-Order Energies.}

\begin{table}[t]
    \small
    \centering
    \begin{tabular}{ll|c|c|c|}
\cline{3-5}
&  & POS & NER & CCG\\ \hline
\multicolumn{2}{|l|}{Linear Chain}  & 89.5 & 90.6 & 92.8 \\
\hline \hline
\multicolumn{1}{|l|}{\multirow{3}{*}{VKP}} & $M=2$ & \textbf{89.9} & 91.1 & \textbf{93.1}\\
\multicolumn{1}{|l|}{} & $M=3$ & 89.8  & \textbf{91.2} & 92.9\\
\multicolumn{1}{|l|}{} & $M=4$ & 89.5  & 90.8 & 92.8\\
\hline \hline
\multicolumn{1}{|l|}{} & $M=1$ & 89.7 & 91.1 & 93.0 \\
\multicolumn{1}{|l|}{\multirow{2}{*}{CNN}} & $M=2$ & \textbf{90.0}  & \textbf{91.3} & \textbf{93.0} \\
\multicolumn{1}{|l|}{} & $M=3$ & 89.9  & 91.2 & 92.9\\
\multicolumn{1}{|l|}{} & $M=4$ & 89.7   & 91.0 & \textbf{93.0} \\
\hline \hline
\multicolumn{1}{|l|}{} & $M=2$ & 89.7 &  90.8 & 92.4\\
\multicolumn{1}{|l|}{\multirow{2}{*}{TLM}} & $M=3$ & 89.8  & 91.0 & 92.7\\
\multicolumn{1}{|l|}{} & $M=4$ & 89.8  & 91.3 & 92.7\\
\multicolumn{1}{|l|}{} & all &  \textbf{90.0}  & \textbf{91.4} & \textbf{92.9}\\
\hline \hline
\multicolumn{1}{|l|}{} & $M=2$ & 89.7 & 90.7 & 92.6  \\
\multicolumn{1}{|l|}{} & $M=4$ & 89.8 & 90.8 & 92.8\\
\multicolumn{1}{|l|}{\multirow{1}{*}{S-Att}} & $M=6$ & \textbf{89.9}  & 90.9 & 92.8 \\
\multicolumn{1}{|l|}{} & $M=8$ &  \textbf{89.9} & \textbf{91.0} & 93.0\\
\multicolumn{1}{|l|}{} & all & 89.7 & 90.8 & \textbf{93.1} \\
\hline
    \end{tabular}
    \caption{Development results for different parameterizations of high-order energies when increasing the window size $M$ of  consecutive labels, where ``all'' denotes the whole relaxed label sequence. The inference network architecture is a one-layer BiLSTM. We ran $t$-tests for the mean performance (over five runs) of our proposed energies (the settings in bold) and the linear-chain energy. 
    All differences are significant at $p<0.001$ for NER and $p<0.005$ for other tasks.
    }
    \label{tab:HighOrder}
\end{table}

We first compare several choices for energy functions within our inference network learning framework. In Section~\ref{sec:hoEnergy}, we considered several ways to define the high-order energy function $F$. We compare performance of the parameterizations on three tasks: POS, NER, and CCG.
The results are shown in Table~\ref{tab:HighOrder}. 

For $\product$ high-order energies, there are small differences between $2$nd and $3$rd order models, however, $4$th order models are consistently worse.  
The CNN high-order energy is best when $M$=2 for the three tasks. Increasing $M$ does not consistently help. 
The tag language model (TLM) works best when scoring the entire label sequence. In the following experiment with TLM energies, we always use it with this ``all'' setting. 
Self-attention (S-Att) also shows better performance with larger $M$. However, the results for NER are not as high overall as for other energy terms. 

Overall, there is no clear winner among the four types of parameterizations, indicating that a variety of high-order energy terms can work well on these tasks, once appropriate window sizes are chosen. We do note differences among tasks: NER benefits more from larger window sizes than POS.

\paragraph{Comparing Structured Energy Terms.}

Above we compared parameterizations of the high-order energy terms. In Table~\ref{ta:main_result}, we compare instantiations of the structured energy term $E_{W}(\y)$: linear-chain energies, skip-chain energies, high-order energies, and fully-connected energies.\footnote{$M$ values are tuned based on dev sets. Tuned $M$ values for POS/NER/CCG/SRL: Skip-Chain: 3/4/3/3; VKP: 2/3/2/2; CNN: 2/2/2/2; TLM: whole sequence; S-Att: 8/8/8/8.}  
We also compare to local classifiers (BiLSTM). The models with structured energies typically improve over the local classifiers, even with just the linear chain energy. 

The richer energy terms tend to perform better than linear chain, at least for most tasks and energies. 
The skip-chain energies benefit from relatively large $M$ values, i.e., 3 or 4 depending on the task. These tend to be larger than the optimal VKP $M$ values. 
We note that S-Att high-order energies work well on SRL. This points to the  benefits of self-attention on SRL, which has been found in recent work \citep{tan2018,strubell-etal-2018-linguistically}.

Both the skip-chain and high-order energy models achieve substantial improvements over the linear chain CRF, notably a gain of 0.8 F1 for NER. The fully-connected energy is not as strong as the others, possibly due to the energies from label pairs spanning a long range. These long-range energies do not appear helpful for these tasks. 

\begin{table}[t]
\setlength{\tabcolsep}{5pt}
\small
\centering
\begin{tabular}{|ll|c|c|c|cc|}
\cline{3-7}
\multicolumn{2}{l|}{}  & \multicolumn{1}{c|}{\multirow{2}{*}{POS}} & \multicolumn{1}{c|}{\multirow{2}{*}{NER}} & \multicolumn{1}{c|}{\multirow{2}{*}{CCG}} & \multicolumn{2}{c|}{SRL}\\ 
\multicolumn{2}{l|}{}   &  &  &  & WSJ & Brown \\
\hline
\multicolumn{2}{|l|}{BiLSTM} &  88.7 &  85.3  & 92.8 &  81.8 & 71.8 \\
\hline \hline
\multicolumn{2}{|l|}{Linear Chain}  & 89.7 & 85.9 & 93.0 & 81.7 & 72.0\\
\hline \hline
\multicolumn{2}{|l|}{Skip-Chain} & 90.0 & \textbf{86.7}  & \textbf{93.3} & 82.1 & \textbf{72.4}\\
\hline \hline
 & VKP & \textbf{90.1}  & \textbf{86.7} & \textbf{93.3} &  81.8 & 72.0\\
 High-& CNN & \textbf{90.1}  & 86.5 & 93.2 & 81.9  & 72.2 \\
 Order & TLM &  90.0 & 86.6 & 93.0 & 81.8  & 72.1\\
 & S-Att & \textbf{90.1}  & 86.5 & \textbf{93.3} & \textbf{82.2}  & 72.2 \\
\hline \hline
\multicolumn{2}{|l|}{Fully-Connected} & 89.8 & 86.3 & 92.9 & 81.4 & 71.4\\
\hline
\end{tabular}
\caption{Test results on all tasks for local classifiers (BiLSTM) and different structured energy functions. 
POS/CCG use accuracy while NER/SRL use F1. 
The architecture of inference networks is one-layer BiLSTM.  
More results are shown in the appendix. 
}
\label{ta:main_result}
\end{table}

\paragraph{Comparison using Deeper Inference Networks.}

\begin{table}[t]
\centering
\small
\begin{tabular}{|ll|c|c|c|}
\cline{3-5}
\multicolumn{2}{l|}{} & \multicolumn{1}{c|}{POS} & \multicolumn{1}{c|}{NER}  & \multicolumn{1}{c|}{CCG} \\ \hline
\multicolumn{2}{|l|}{2-layer BiLSTM} & 88.8  & 86.0 & 93.4 \\
\hline \hline
\multicolumn{2}{|l|}{BiLSTM-CRF}  & 89.2  & 87.3 & 93.1  \\
\hline \hline
\multicolumn{2}{|l|}{Linear Chain}  & 90.0 & 86.6 &  93.7 \\
\hline \hline
\multicolumn{2}{|l|}{Skip-Chain} & \textbf{90.2} &  \textbf{87.5} & \textbf{93.8} \\
\hline \hline
& VKP  & \textbf{90.2}  & 87.2 &  \textbf{93.8} \\
High-& CNN  & \textbf{90.2}  & 87.3 &  93.6 \\
Order & TLM  & 90.1 & 87.1 &  93.6 \\
& S-Att & 90.0 & 87.3 &  93.7 \\
\hline \hline
\multicolumn{2}{|l|}{Fully-Connected} & 90.0 & 87.2 &  93.3 \\
\hline
    \end{tabular}
    \caption{Test results when inference networks have 2 layers (so the local classifier baseline also has 2 layers).}
    \label{tab:2BiLSTM}
\end{table}

Table~\ref{tab:2BiLSTM} compares methods when using 2-layer BiLSTMs as inference networks.\footnote{$M$ values are retuned based on dev sets when using 2-layer inference networks. Tuned $M$ values for POS/NER/CCG: Skip-Chain: 3/4/3; VKP: 2/3/2; CNN: 2/2/2; TLM: whole sequence; S-Att: 8/8/8.} 
The deeper inference networks reach higher performance across all tasks compared to 1-layer  inference networks. 

We observe that inference networks trained with skip-chain energies and high-order energies achieve better results than BiLSTM-CRF on the three datasets (the Viterbi algorithm is used for exact inference for BiLSTM-CRF). This indicates that adding richer energy terms can make up for approximate inference during training and inference. Moreover, a 2-layer BiLSTM is much cheaper computationally than Viterbi, especially for tasks with large label sets. 

\subsection{Results on Noisy Datasets}

\begin{table}[t]
    \small
    \centering
    \begin{tabular}{|l|c|c|c|}
\cline{2-4}
\multicolumn{1}{l|}{} & \multicolumn{1}{c|}{$\alpha$=0.1} & \multicolumn{1}{c|}{$\alpha$=0.2} & \multicolumn{1}{c|}{$\alpha$=0.3}  \\ \hline
BiLSTM &  75.0  & 67.2 & 58.8 \\
\hline \hline
Linear Chain & 75.2  & 67.4 &  59.1\\
\hline \hline
Skip-Chain ($M$=4) &  75.5  & 67.9 &  59.5\\
\hline \hline
VKP ($M$=3) &  75.3  & 67.7 &  59.3\\
CNN ($M$=0)  & 75.7 & 67.9 & 59.4 \\  
CNN ($M$=2)  & 76.3  & 68.6  & 60.2\\
CNN ($M$=4)  & \textbf{76.7}  & \textbf{69.8} &  \textbf{60.4}\\
TLM  & 76.0  & 67.8 &  59.9\\
S-Att ($M$=8)  & 75.6  & 67.6 &  59.7\\\hline
    \end{tabular}
    \caption{UnkTest setting for NER: 
    words in the test set are replaced by the unknown word symbol with probability $\alpha$. For CNN energies (the settings in bold) and linear-chain energy, they differ significantly with $p<0.001$.} 
    \label{tab:unktest}
\end{table}

\begin{table}[t]
    \small
    \centering
    \begin{tabular}{|l|c|c|c|}
\cline{2-4}
\cline{2-4}
\multicolumn{1}{l|}{} & \multicolumn{1}{c|}{$\alpha$=0.1} & \multicolumn{1}{c|}{$\alpha$=0.2}  & \multicolumn{1}{c|}{$\alpha$=0.3} \\ \hline
BiLSTM &  80.1 &  76.0 & 70.6\\
\hline \hline
Linear Chain &  80.4 & 76.3  & 70.9\\
\hline \hline
Skip-Chain ($M$=4) &  81.2 &  76.7 &  71.2\\
\hline \hline
VKP ($M$=3) &  81.4  & 76.8 &  71.4\\
CNN ($M$=0) & 81.1 & 76.7 &  71.5\\
CNN ($M$=2) & 81.8 & 77.0 &  \textbf{71.8}\\
CNN ($M$=4)  & \textbf{82.0} & \textbf{77.1} &  71.7\\
TLM &  80.9 &  76.3 &  71.1\\
S-Att ($M$=8) & 81.4  & 76.9 &  71.4\\\hline
    \end{tabular}
    \caption{UnkTrain setting for NER: training on noisy text, evaluating on noisy test sets. Words are replaced by the unknown word symbol with probability $\alpha$. For CNN energies (the settings in bold) and linear-chain energy, they differ significantly with $p<0.001$. }
    \label{tab:unktrain}
\end{table}

We now consider the impact of our structured energy terms in noisy data settings. Our motivation for these experiments stems from the assumption that structured energies will be more helpful when there is a weaker relationship between the observations and the labels. One way to achieve this is by introducing noise into the observations. 

So, we create new datasets: for any given sentence, we randomly replace a token $x$ with an unknown word symbol ``UNK'' with probability $\alpha$. 
From previous results, we see that NER shows more benefit from structured energies, so we focus on NER 
and consider two 
settings: 
    \textbf{UnkTest}: train on clean text, evaluate on noisy text; and 
  \textbf{UnkTrain}: train on noisy text, evaluate on noisy text.
  
Table~\ref{tab:unktest} shows results for UnkTest. 
CNN 
energies are best among all structured energy terms, including the different parameterizations. 
Increasing $M$ improves F1, showing that high-order information helps the model recover from the high degree of noise. 
Table~\ref{tab:unktrain} shows results for UnkTrain. 
The CNN high-order energies again yield large gains: roughly 2 points compared to the local classifier and 1.8 compared to the linear chain energy.

\section{Incorporating BERT} 
Researchers have recently been applying large-scale pretrained transformers like BERT~\citep{devlin-etal-2019-bert} to many tasks, including sequence labeling. 
To explore the impact of high-order energies on BERT-like models, we now consider experiments that use BERT$_{\text{BASE}}$ in various ways. We use two baselines: (1) BERT finetuned for NER using a local loss, and (2) a CRF using BERT features (``BERT-CRF''). 
Within our framework, we also experiment with using BERT in both the energy function and inference network architecture. 
That is, the ``input feature vector'' in Equation~\ref{eqn:energy} is replaced by the features from BERT. The energy and inference networks are trained with the objective in Section~\ref{sec:joint}. For the training of energy function and inference networks, we use Adam with learning rate $5\mathrm{e}\!-\!5$, a batch size of 32, and L2 weight decay of $1\mathrm{e}\!-\!5$. The results are shown in Table~\ref{tab:berttest}.\footnote{Various high-order energies were explored. We found the skip-chain energy ($M$=3) to achieve the best performance (96.28) on the dev set, so we use it when reporting the test results.}

There is a slight improvement when moving from BERT trained with the local loss to using BERT within the CRF (92.13 to 92.34). There is little difference (92.13 vs.~92.14) between the locally-trained BERT model and when using the linear-chain energy function within our framework. However, when using the higher-order energies, the difference is larger (92.13 to 92.46).

\begin{table}[t]
    \small
    \centering
    \begin{tabular}{lc|c|}
\multicolumn{3}{l}{\textbf{Baselines:}}\\ 
\hline
\multicolumn{2}{|l|}{BERT (local loss)} &  92.13  \\

\multicolumn{2}{|l|}{BERT-CRF} &   92.34 \\
\hline 
\multicolumn{3}{l}{\textbf{Energy-based inference networks:}} \\
\hline
\multicolumn{2}{|l|}{Linear Chain }   & 92.14  \\
\multicolumn{2}{|l|}{Skip-Chain ($M$=3)} & 92.46  \\
\hline
    \end{tabular}
    \caption{Test results for NER when using BERT. When using energy-based inference networks (our framework), BERT is used in both the energy function and as the inference network architecture.} 
    \label{tab:berttest}
\end{table}

\section{Analysis of Learned Energies} 

In this section, we visualize our learned energy  functions for NER to see what structural dependencies among labels have been captured. 

\begin{figure}[t]
\small
\begin{center}
\includegraphics[width=1\linewidth]{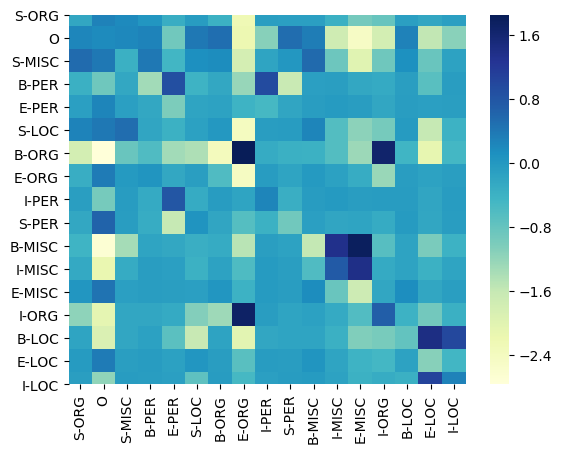}
\\(a) Skip-chain energy matrix $W_1$.
\includegraphics[width=1\linewidth]{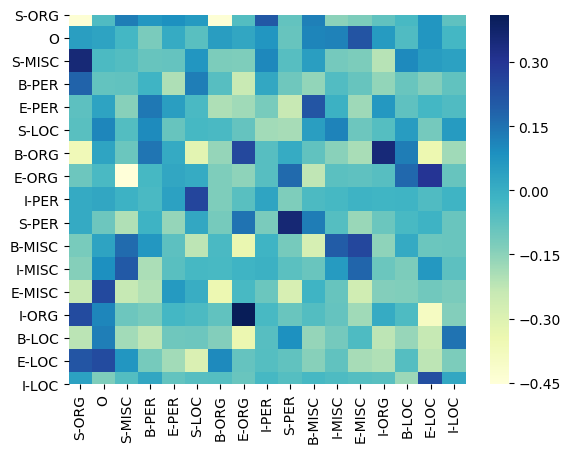}
\\(b) Skip-chain energy matrix $W_3$.
\end{center}
\caption{Learned pairwise potential matrices $W_1$ and $W_3$ for NER with skip-chain energy. The rows correspond to earlier labels and the columns correspond to subsequent labels.
}
\label{fig:skip_chain}
\end{figure}

\begin{figure}[h!]
\small
\begin{center}
\includegraphics[width=1\linewidth]{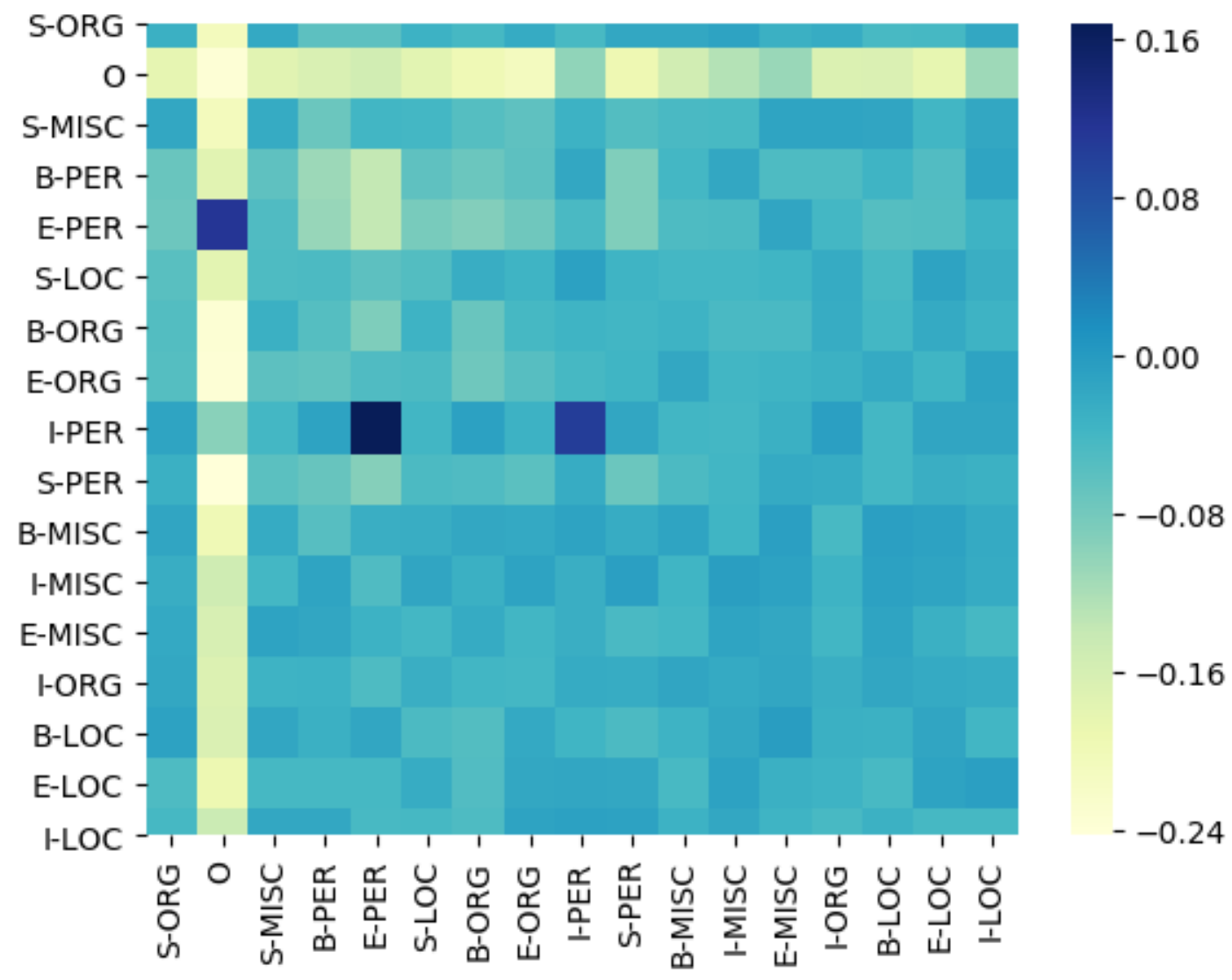}
\end{center}
\caption{Learned 2nd-order VKP energy matrix beginning with B-PER in NER dataset.}
\label{fig:kp}
\end{figure}

Figure~\ref{fig:skip_chain} visualizes two matrices in the skip-chain energy with $M=3$. We can see strong associations among labels in neighborhoods from $W_1$. For example, B-ORG and I-ORG are more likely to be followed by E-ORG. The $W_3$ matrix shows a strong association between I-ORG and E-ORG, which implies that the length of organization names is often long in the dataset. 


\begin{table}[t]
\centering
\small
\begin{tabular}{llll}
filter 26 & B-MISC & I-MISC & E-MISC \\
filter 12 & B-LOC & I-LOC & E-LOC \\
filter 15 & B-PER & I-PER & I-PER \\
filter 5 & B-MISC &  E-MISC & O \\
filter 6 & O & B-LOC & I-LOC \\
filter 16 & S-LOC & B-ORG & I-ORG \\
filter 44 & B-PER & I-PER & I-PER \\
filter 3 & B-MISC &  I-MISC & E-MISC \\
filter 2 & I-LOC &  E-LOC & O \\
filter 45 & O & B-LOC & E-LOC \\
\end{tabular}
\caption{Top 10 CNN filters with high inner product with 3 consecutive labels for NER. 
}\label{ta:cnnVis}
\end{table}

For the VKP energy with $M$=3, Figure~\ref{fig:kp} shows the learned matrix when the first label is B-PER, showing that B-PER is likely to be  followed by ``I-PER E-PER'', ``E-PER O'', or ``I-PER I-PER''. 

In order to visualize the learned CNN filters, we calculate the inner product between the filter weights and consecutive labels. For each filter, we select the sequence of consecutive labels with the highest inner product. Table~\ref{ta:cnnVis} shows the 10 filters with the highest inner product and the corresponding label trigram. 
All filters give high scores for structured label sequences with a strong local dependency, such as ``B-MISC {} I-MISC {} E-MISC" and ``B-LOC {}  I-LOC {} E-LOC", etc. 
Figure~\ref{fig:cnn} in the appendix shows these inner product scores of 50 CNN filters on a sampled NER label sequence. We can observe that filters learn the sparse set of label trigrams with strong local dependency.

\section{Conclusion}
We explore arbitrary-order models with different neural parameterizations on sequence labeling tasks via energy-based inference networks. This approach achieve substantial improvement using high-order energy terms, especially in noisy data conditions, while having same decoding speed as simple local classifiers. 

\section*{Acknowledgments}
We would like to thank Sam Wiseman for help regarding high-order CRFs and the reviewers for insightful comments. 
This research was supported in part by an Amazon Research Award to K.~Gimpel.

\bibliographystyle{acl_natbib}
\bibliography{anthology,emnlp2020}

\appendix
\section{Appendices}
\label{sec:appendix}

\begin{table*}[t!]
\centering
\small
\begin{tabular}{|l|cc|cc|cc|ccc|}
\cline{2-10}
\multicolumn{1}{l|}{}  & \multicolumn{2}{c|}{POS} & \multicolumn{2}{c|}{NER} & \multicolumn{2}{c|}{CCG} & \multicolumn{3}{c|}{SRL}\\ 
\multicolumn{1}{l|}{} & Dev & Test & Dev & Test & Dev & Test & Dev & WSJ & Brown \\
\hline
BiLSTM & 88.6 & 88.7 & 90.4 & 85.3 & 92.6 & 92.8 & 80.2 & 81.8 & 71.8\\
BiLSTM + CRF  & 89.1 & 89.2 & 91.6 & 87.3 & 93.0 & 93.1 & - & - & - \\
\hline
Linear Chain & 89.5 & 89.7 & 90.6 & 85.9 & 92.8 & 93.0 & 80.3 & 81.7 & 72.0\\
VKP (M=2) & 89.9 & 90.1 & 91.1 & 86.5 & 93.1 & 93.3 & 80.3 & 81.8 & 72.0\\
VKP (M=3) & 89.7 & 89.8 & 91.2 & 86.7 & 92.9 & 93.0 & 80.1	& 81.6 & 71.6\\
VKP (M=4) & 89.4 & 89.5 & 90.8 & 86.3 & 92.8 & 93.0 & 79.9 & 81.2 & 71.3\\
VKP (M=[2,3,4]) & 89.8 & 89.9 & 91.0 & 86.5 & 93.0 & 93.3 & 80.3 & 81.9 & 71.8\\
Skip Chain (M=2) & 89.7 & 89.8 & 90.8 & 86.2 & 92.8 & 93.1 & 80.3 &	81.8 & 71.8\\
Skip Chain (M=3) & 89.9	& 90.0 & 91.2 & 86.5 & 93.0 & 93.3 & 80.4 & 82.1 & 72.4\\
Skip Chain (M=4) & 89.8	& 89.9 & 91.3 & 86.7 & 92.7 & 92.8 & 80.2 & 81.6 & 71.7\\
Skip-Chain (M=5) & 89.5 & 89.6 & 91.0 & 86.2 & 92.5 & 92.7 & 80.2 & 81.7 & 71.7\\
Fully Connect (M=20) & 89.7 & 89.8 & 91.1 & 86.3 & 92.8 & 92.9 & 80.0 & 81.4 & 71.4\\
CNN (M=0) & 89.6 & 89.8 & 90.9 & 86.2 & 92.6 & 92.8 & 80.0 & 81.7 & 71.8\\
CNN (M=1) & 89.7 & 89.8 & 91.1 & 86.4 & 92.8 & 93.0 & 80.1 & 81.8 & 72.0\\
CNN (M=2) & 90.0 & 90.1 & 91.3 & 86.5 & 93.0 & 93.2 & 80.3 & 81.9 & 72.2\\
CNN (M=3) & 89.9 & 89.9 & 91.2 & 86.4 & 92.9 & 93.0 & 80.0 & 81.7 & 71.9\\
CNN (M=4) & 89.7 & 89.8 & 91.0 & 86.2 & 93.0 & 93.1 & 80.2 & 81.7 & 72.2\\
CNN (M=1,2,3) & 90.0 & 90.0 & 91.3 & 86.6 & 93.1 & 93.3 & 80.3 & 82.0 & 72.2\\
TLM (M=1) & 89.6 & 89.7 & 90.9 & 86.3 & 92.4 & 92.6 & 79.8 & 81.3 & 71.3\\
TLM (M=2) & 89.7 & 89.8 & 90.8 & 86.3 & 92.4 & 92.7 & 80.0 & 81.6 & 71.7\\
TLM (M=3) & 89.8 & 89.8 & 91.0 & 86.4 & 92.7 & 92.9 & 80.1 & 81.7 & 71.9\\
TLM (M=4) & 89.8 & 90.0 & 91.3 & 86.5 & 92.7 & 92.8 & 80.0 & 81.6 & 71.8\\
TLM & 90.0 & 90.0 & 91.4 & 86.6 & 92.9 & 93.0 & 80.2 & 81.8 & 72.1\\
S-Att(M=2) & 89.7 & 89.8 & 90.7 & 86.3 & 92.6 & 92.8 & 80.0 & 81.6 & 71.8\\
S-Att(M=4) & 89.8 & 89.9 & 90.8 & 86.4 & 92.8 & 93.0 & 80.0 & 81.7 & 71.8\\
S-Att(M=6) & 89.9 & 90.0 & 90.9 & 86.4 & 92.8 & 93.1 & 80.2 & 81.9 & 72.0\\
S-Att(M=8) & 89.9 & 90.1 & 91.0 & 86.5 & 93.0 & 93.3 & 80.4 & 82.2 & 72.2\\
S-Att & 89.7 & 89.9 & 90.8 & 86.4 & 93.1 & 93.3 & 80.3 & 82.0 & 72.2\\
\hline
\end{tabular}
\caption{Results on all tasks for local classifiers and different structured energy functions: linear-chain energy, Kronecker Product high-order energies, skip-chain energy and fully-connected energies. The metrics of the four tasks POS, NER, CCG, SRL are accuracy, F1, accuracy and F1. The architecture of inference networks is one-layer BiLSTM.}
\label{ta:result}
\end{table*}

\begin{table*}[t!]
\centering
\small
\begin{tabular}{|l|cc|cc|cc|}
\cline{2-7}
\multicolumn{1}{l|}{} & \multicolumn{2}{c|}{POS} & \multicolumn{2}{c|}{NER}  & \multicolumn{2}{c|}{CCG} \\
\multicolumn{1}{l|}{} & Dev & Test & Dev & Test & Dev & Test \\\hline
2-layer BiLSTM & 88.7 & 88.8 & 90.9 & 86.0 & 93.2 & 93.4\\
Linear Chain & 89.9 & 90.0 & 91.2 & 86.6 & 93.3 & 93.7\\
Skip-Chain & 90.0 & 90.2 & 91.7 & 87.5 & 93.5 & 93.8\\
VKP & 89.9 & 90.2 & 91.5 & 87.2 & 93.6 & 93.8\\
CNN & 90.0 & 90.2 & 91.5 & 87.3 & 93.5 & 93.6\\
TLM & 89.9 & 90.1 & 91.4 & 87.1 & 93.3 & 93.6\\
S-Att (M=8) & 89.9 & 90.0 & 91.6 & 87.3 & 93.5 & 93.7 \\
Fully Connected & 89.8 & 90.0 & 91.4 & 87.2 & 93.2 & 93.3\\\hline
    \end{tabular}
    \caption{Results when inference networks use 2-layer BiLSTMs (so the local classifier baseline also has 2 layers).}
    \label{tab:TwoBiLSTM}
\end{table*}

\begin{table*}[t!]
\small
    \centering
    \begin{tabular}{|l|cc|cc|cc|}
\cline{2-7}
\multicolumn{1}{l|}{} & \multicolumn{2}{c|}{$\alpha$=0.1} & \multicolumn{2}{c|}{$\alpha$=0.2} & \multicolumn{2}{c|}{$\alpha$=0.3}  \\
\multicolumn{1}{l|}{} & Dev & Test & Dev & Test & Dev & Test \\\hline
BiLSTM & 80.0 & 75.0 & 70.1 & 67.2 & 62.4 & 58.8 \\
Linear Chain & 80.2 & 75.2 & 70.3 & 67.4 & 62.7 & 59.1\\
Skip-Chain (M=4) & 80.6 & 75.5 & 70.9 & 67.9 & 63.2 & 59.5\\
VKP (M=3) & 80.5 & 75.3 & 70.5 & 67.7 & 62.8 & 59.3\\
CNN (M=0) & 80.8 & 75.7 & 71.3 & 67.9 & 63.3 & 59.4 \\  
CNN (M=2) & 81.4 & 76.3 & 72.4 & 68.6 & 64.0 & 60.2\\
CNN (M=4) & 81.9 & 76.7 & 73.0 & 69.8 & 64.5 & 60.4\\
TLM & 81.0 & 76.0 & 71.3 & 67.8 & 63.8 & 59.9\\
S-Att (M=8) & 80.6 & 75.6 & 71.5 & 67.6 & 63.2 & 59.7\\\hline
    \end{tabular}
    \caption{UnkTest setting for NER: 
    Words in the test set are randomly replaced by the unknown word symbol with probability $\alpha$.}
    \label{tab:noiseInput}
\end{table*}

\begin{table*}[h!]
    \centering
    \small
    \begin{tabular}{|l|cc|cc|cc|}
\cline{2-7}
\multicolumn{1}{l|}{} & \multicolumn{2}{c|}{$\alpha$=0.1} & \multicolumn{2}{c|}{$\alpha$=0.2}  & \multicolumn{2}{c|}{$\alpha$=0.3} \\
\multicolumn{1}{l|}{} & Dev & Test & Dev & Test & Dev & Test \\\hline
BiLSTM & 85.0 & 80.1 & 80.0 & 76.0 & 75.0 & 70.6\\
Linear Chain & 85.4 & 80.4 & 80.5 & 76.3 & 75.2 & 70.9\\
Skip-Chain (M=4) & 85.7 & 81.2 & 80.7 & 76.7 & 75.4 & 71.2\\
VKP (M=3) & 85.9 & 81.4 & 81.0 & 76.8 & 75.5 & 71.4\\
CNN (M=0) & 85.6 & 81.1 & 80.8 & 76.7 & 75.6 & 71.5\\
CNN (M=2) & 86.0 & 81.8 & 81.2 & 77.0 & 76.1 & 71.8\\
CNN (M=4) & 86.1 & 82.0 & 81.2 & 77.1 & 75.9 & 71.7\\
TLM & 85.6 & 80.9 & 80.6 & 76.3 & 75.3 & 71.1\\
S-Att (M=8) & 85.8 & 81.4 & 81.0 & 76.9 & 75.6 & 71.4\\\hline
    \end{tabular}
    \caption{UnkTrain setting for NER: training on noisy text, evaluating on noisy test sets. Words are randomly replaced by the unknown word symbol with probability $\alpha$.}
    \label{tab:noiseInput1}
\end{table*}

\begin{figure*}[h!]
\small
\begin{center}
\includegraphics[width=\linewidth]{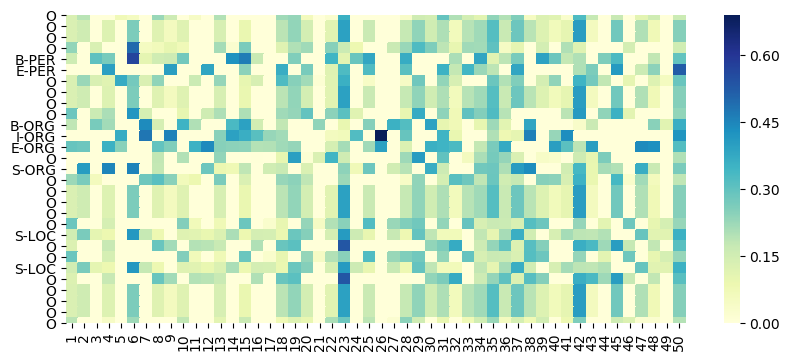}
\end{center}
\caption{Visualization of the scores of 50 CNN filters on a sampled label sequence. We can observe that filters learn the sparse set of label trigrams with strong local dependency. 
}
\label{fig:cnn}
\end{figure*}
%

\end{document}